\documentclass[10pt,twocolumn,letterpaper]{article}

\usepackage[pagenumbers]{cvpr}

\definecolor{cvprblue}{rgb}{0.21,0.49,0.74}
\usepackage[breaklinks,colorlinks,allcolors=cvprblue]{hyperref}

\usepackage{pifont}
\newcommand{\cmark}{\ding{51}}
\newcommand{\xmark}{\ding{55}}

\def\confName{CVPR}
\def\confYear{2027}

\definecolor{gaincolor}{RGB}{185,85,35}
\newcommand{\gain}[1]{{%
  \fontsize{6.8}{8}\selectfont\color{gaincolor}#1}}

\title{AdaGEM: Granularity Adaptation and Geospatial Evidence Modulation for Ultra-High-Resolution Remote Sensing Understanding}

\author{
Kexin Ma\textsuperscript{1}\quad
Jing Xiao\textsuperscript{1}\quad
Bowen Xing\textsuperscript{1}\quad
Liang Liao\textsuperscript{2}\quad
Chia-Wen Lin\textsuperscript{3}\\[3pt]
\textsuperscript{1}School of Artificial Intelligence, Wuhan University\\
\textsuperscript{2}Hangzhou Institute of Technology, Xidian University\\
\textsuperscript{3}Department of Electrical Engineering, National Tsing Hua University
}

\hypersetup{
  pdftitle={AdaGEM: Granularity Adaptation and Geospatial Evidence Modulation for Ultra-High-Resolution Remote Sensing Understanding},
  pdfauthor={Kexin Ma, Jing Xiao, Bowen Xing, Liang Liao, Chia-Wen Lin},
  pdfsubject={}
}

\usepackage{amsmath}
\usepackage{amssymb}
\usepackage{xcolor}
\definecolor{lightgray}{gray}{1}
\usepackage{colortbl}
\usepackage{booktabs}
\definecolor{lightblue}{RGB}{224,232,240}
\usepackage{multirow}
\usepackage{subcaption}
\usepackage{graphicx}
\usepackage[table]{xcolor}

\definecolor{sectionbg}{RGB}{245,246,247} 
\definecolor{oursbg}{RGB}{232,242,250}

\begin{document}
\maketitle
\begin{abstract}
Ultra-high-resolution (UHR) remote sensing imagery provides rich spatial details and extensive geographic coverage for Earth observation understanding. However, such imagery generates quadratically growing visual token sequences, where sparse task-critical evidence is easily overwhelmed by redundant and task-irrelevant tokens, leading to excessive computation and ineffective evidence preservation. Existing methods seek to balance computational efficiency and task effectiveness through selective multi-scale inspection or direct token pruning, yet often overlook task-dependent granularity requirements and struggle to distinguish sparse critical evidence from overwhelming visual redundancy. To address these challenges, we propose AdaGEM, a unified framework that jointly adapts visual granularity to task demands and modulates token selection to preserve essential geospatial evidence. Specifically, a Granularity Router predicts detail requirements and evidence scope from the query, then selects input resolution and token budget via a rule-based mapping. At the adapted granularity, a Geospatial Evidence Modulator selects complementary visual tokens to preserve critical evidence and spatial context while suppressing redundancy in homogeneous regions. Extensive experiments across diverse remote sensing benchmarks and multiple MLLM backbones demonstrate that AdaGEM consistently improves task performance over strong baselines and existing UHR methods while substantially reducing inference costs.
\end{abstract}

\section{Introduction}
Multimodal Large Language Models (MLLMs) have demonstrated strong capabilities in remote sensing understanding, supporting applications such as scene interpretation \cite{kuckreja2023geochat,10.1007/978-3-031-72904-1_26,skymoe} and geospatial question answering \cite{EarthDial, Zhu_2025_skysenseo, LIU2026113717, danish2025geobenchvlm}. Recent progress in satellite and aerial imaging \cite{denis2026pleiades,farella2025ultracam, rshr_benchmark} has made Ultra-High-Resolution (UHR) imagery increasingly available, creating new opportunities for MLLMs to advance Earth observation. However, processing such imagery incurs substantial computational costs, as visual token counts grow quadratically with linear input resolution. This inefficiency is further exacerbated by the uneven spatial distribution of visual information, with extensive homogeneous regions coexisting with sparsely distributed small objects and complex spatial structures. Consequently, a substantial portion of the computation tends to be spent on processing redundant visual tokens, potentially at the expense of capturing key objects and fine-grained spatial structures essential for accurate understanding. 

\begin{figure}
\centering
\includegraphics[width=\linewidth]{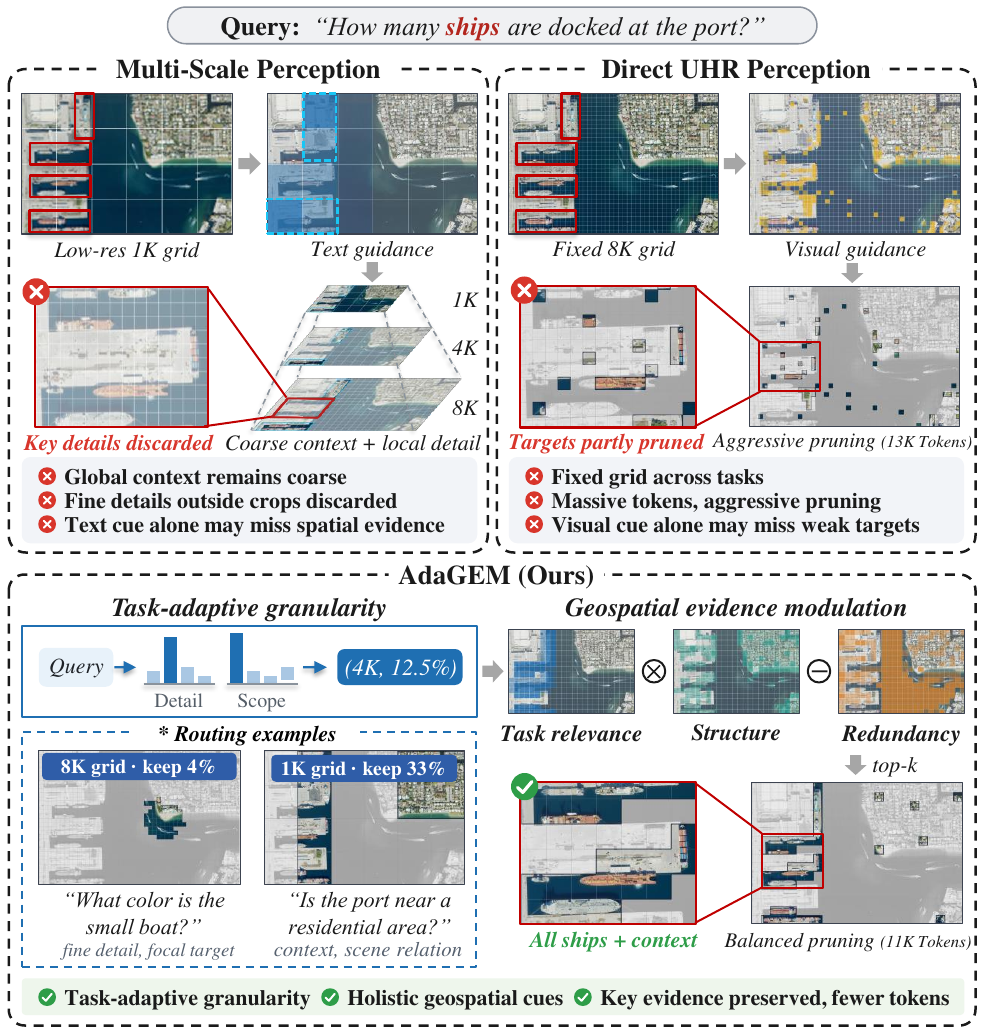}
\vspace{-5mm}
\caption{\textbf{Perception paradigms for UHR RS understanding.} Multi-scale perception leaves global context coarse, while direct UHR perception requires aggressive pruning. Text or visual guidance alone may overlook complementary evidence. AdaGEM combines task-adaptive granularity with complementary geospatial cues to preserve key evidence with fewer tokens.} \label{fig0}
\vspace{-4mm}
\end{figure}

Building on advances in token pruning \cite{yang2025visionzip, zhang2024sparsevlm} and region selection \cite{li2024mini, shen2025zoomeye, wang2025rap} for general-purpose MLLMs \cite{liu2024llava, bai2025qwen25vltechnicalreport,llava_uhd}, recent works on remote sensing (RS) have explored two practical approaches for efficient UHR understanding. Multi-scale perception methods \cite{LRS-VQA,ZoomSearch,zoomearth,WeaveEarth} concentrate UHR processing on selected image regions to capture task-relevant visual details. GeoLLaVA-\textit{8K} \cite{wang2025geollava8k} further enables direct UHR perception of images up to 8K resolution by reducing background redundancy through token merging \cite{bolya2022tome} and preserving informative objects through visual attention-based selection \cite{yang2025visionzip}. Despite their effectiveness, existing approaches remain limited in two important respects. First, visual evidence selection based on task relevance or visual attention may overlook weakly activated regions that provide essential spatial or contextual cues. Prior studies have also identified biases in these selection signals \cite{ma2026apet,chen2025variation}, which may direct processing toward task-irrelevant regions. Second, multi-scale perception methods leave fine details in unselected regions underexplored, whereas fixed UHR inputs generate massive fine-grained token sequences that require aggressive pruning. Under these fixed perception paradigms, important geospatial evidence may be lost or represented only at a coarse global level, ultimately limiting UHR RS understanding performance.

To address these limitations, we revisit efficient UHR remote sensing understanding from two complementary perspectives: holistic geospatial evidence preservation and task-adaptive visual granularity. We first consider key characteristics of UHR remote sensing images (RSIs), including target diversity, spatial dependencies, and homogeneous regions, to develop an evidence modulation strategy that integrates task relevance, geospatial structure, and local redundancy. Structural cues help preserve spatial and contextual information that task relevance alone may underemphasize, while redundancy suppression limits the retention of repetitive visual content. Building on this strategy, we further examine whether higher input resolution and greater token retention consistently improve understanding. Our analysis reveals task-dependent granularity preferences and shows that retaining more tokens at a fixed resolution does not necessarily improve accuracy. These findings motivate jointly adapting input resolution and token retention to task requirements, aiming to provide sufficient visual detail while preserving complementary geospatial evidence.

In this work, we introduce AdaGEM, a framework for efficient UHR remote sensing understanding that jointly adapts task-specific visual granularity and enables holistic geospatial evidence preservation. Specifically, a Granularity Router predicts detail specificity and evidence scope from the input question, which jointly determine the input resolution, token retention ratio, and wavelet configuration. Then a Geospatial Evidence Modulator integrates task relevance, wavelet-based structural cues, and dynamic local redundancy to identify informative and complementary visual tokens, addressing the limitations of semantic relevance alone in capturing structural details. By coordinating these two components, AdaGEM adapts both visual granularity and token selection to different remote sensing tasks. Extensive experiments across benchmarks and backbones demonstrate its effectiveness in improving accuracy while reducing inference costs.

Our main contributions are summarized as follows:
\begin{itemize}
\item We rethink UHR remote sensing understanding as a joint problem of task-adaptive visual granularity and geospatial evidence preservation, aligning visual detail with task demands while preserving sparse task-critical evidence under constrained visual token budgets.

\item We develop a geospatial evidence modulation strategy that jointly accounts for target diversity, spatial dependencies, and regional homogeneity to preserve sparse task-critical evidence and complementary spatial context while suppressing redundant visual content.

\item We propose AdaGEM, a unified framework that coordinates task-driven granularity adaptation with geospatial evidence modulation, jointly adapting input resolution and visual token allocation to diverse task requirements for UHR remote sensing understanding.
\end{itemize}

\begin{figure*}[t]
\centering
\includegraphics[width=0.98\textwidth]{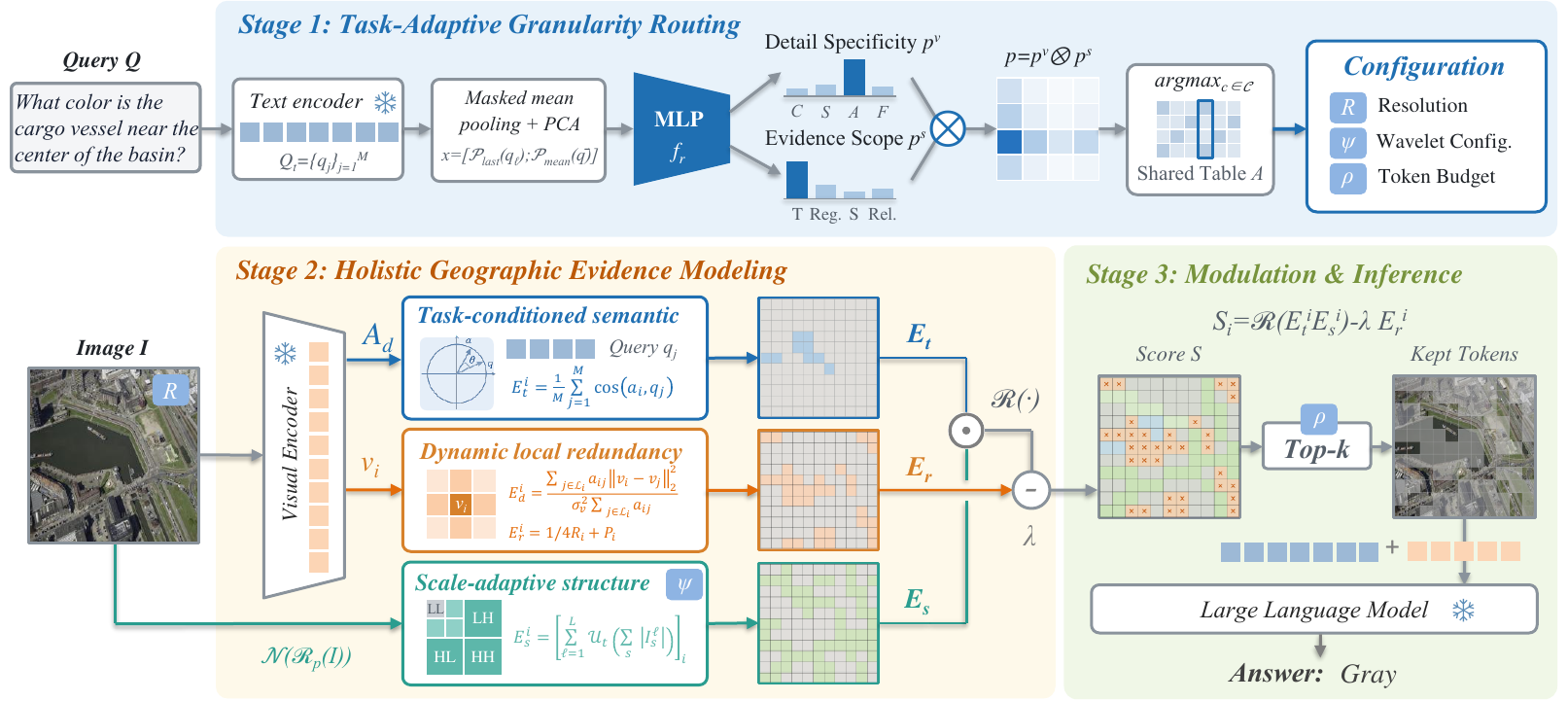}
\vspace{-1mm}
\caption{\textbf{Overview of AdaGEM.} A lightweight router predicts the required visual granularity including detail specificity and evidence scope and selects the corresponding configuration. Wavelet cues, decoupled attention and feature affinity model spatial structure, task relevance, and local redundancy, respectively. Their fusion yields more holistic importance scores for top\text{-}$k$ selection before LLM prefill.} \label{fig_method}
\vspace{-2mm}
\end{figure*}

\section{Related Work}
\subsection{Token Pruning for RS MLLMs}
Visual token pruning improves the computational efficiency of MLLMs by filtering out redundant visual tokens. Early work reduced visual redundancy through token merging \cite{bolya2022tome,wang2025folder, wen-etal-2025-stop}. More recent studies have explored two broad pruning paradigms: task-centric methods exploit attention between textual and visual tokens to retain query-relevant information \cite{zhang2024sparsevlm,xing2024pyramiddrop, chen2025variation}, while vision-centric methods exploit visual attention \cite{dynamicvit,yang2025visionzip,zhangvscan,Shang_2025_ICCV} or feature similarity \cite{ma2026apet,zou2025don,tong2025flowcut}. Recent RS MLLMs extend these strategies to high-resolution (HR) and UHR imagery. RFM-DIP \cite{LRS-VQA} distills text-guided localization from the LLM into a lightweight module, which guides coarse-to-fine tile selection and token pruning within a dynamic image pyramid. GeoLLaVA-\(\textit{8K}\) \cite{wang2025geollava8k} merges semantically similar background tokens and uses visual-attention-anchored selection to retain informative tokens, enabling inputs up to 8K resolution. UHR-BAT \cite{dang2026uhrbat} aligns query-guided importance across scales and combines region-wise token preservation with merging to maintain regional coverage. 

These methods extend natural-image token pruning to remote sensing, yet selection based on task relevance or visual attention alone may overlook fine-grained structural cues and broader spatial context essential for geographic understanding. We therefore develop a more holistic evidence modulation strategy to preserve geospatial evidence at different granularities under limited token budgets.

\subsection{Multi-Scale Perception for RS MLLMs}
Multi-scale perception integrates broad spatial context with fine local details for HR remote sensing image understanding. Beyond processing predefined image tiles, several general-purpose MLLM frameworks selectively inspect regions according to the question. V* \cite{Wu_2024_CVPR_vstar} performs guided visual search to locate relevant objects and combines the retrieved local evidence with a global image representation. RAP \cite{wang2025rap} retrieves question-relevant crops and arranges them according to their spatial relationships, while ZoomEye \cite{shen2025zoomeye} searches an image hierarchy through progressive zooming to identify informative regions. Related strategies have also been developed for HR and UHR remote sensing imagery. EarthDial \cite{EarthDial} adopts adaptive image tiling alongside a global thumbnail, combining local detail and scene context across varying input resolutions. ZoomSearch \cite{ZoomSearch} conducts hierarchical region search and reassembles selected patches according to their spatial layout, whereas ZoomEarth \cite{zoomearth} learns a crop-and-zoom policy to acquire detailed evidence from the original high-resolution image. WeaveEarth \cite{WeaveEarth} combines a global overview, local evidence, and spatial metadata to organize complementary regions while preserving their spatial relationships. 

These methods selectively inspect regions at high resolution to capture task-relevant details and supplement them with low-resolution global context. However, coarse context may still miss fine details outside these regions, limiting tasks that require fine-grained evidence across the scene. We therefore jointly adapt image granularity and token budget to task-specific detail and coverage requirements.

\section{Methodology}
Given a UHR remote sensing image (RSI) $I$ and a query $Q$, we aim to preserve visual evidence for accurate geospatial question answering while improving MLLM efficiency. Since input resolution affects each token's spatial extent, AdaGEM couples task-adaptive granularity routing with holistic geospatial evidence modulation. As shown in Figure \ref{fig_method}, a lightweight router predicts detail specificity and evidence scope to jointly determine the granularity of candidate representations, the token retention budget, and the wavelet configuration for structural analysis. Under this configuration, the modulator integrates task relevance, spatial structure, and dynamic local redundancy to retain complementary visual evidence for MLLM inference.

\subsection{Task-Adaptive Granularity Routing}
To adapt visual granularity to the query, we consider its detail specificity together with the scope of supporting evidence. Since comparable detail may be required for a focal target or across spatially distributed regions, resolution selection must be coordinated with evidence retention. We thus decompose routing into two complementary predictions: detail specificity, spanning context, structure, attributes, and fine-grained cues; and evidence scope, classified as target-specific, regional, scene-wide, or relational.

Given the text-token features $Q_t=\{q_j\}_{j=1}^{M}$ from the text encoder, we obtain the query representation as
\begin{equation}
\begin{aligned}
\bar q &=
\frac{\sum_{j=1}^{M}m_jq_j}{\sum_{j=1}^{M}m_j}, x = \left[
\mathcal{P}_{\mathrm{last}}(q_{\ell});
\mathcal{P}_{\mathrm{mean}}(\bar q)
\right],
\end{aligned}
\end{equation}
where $m_j$ is the attention mask, $q_{\ell}$ is the last valid token feature, and $\mathcal{P}_{\mathrm{last}}$ and $\mathcal{P}_{\mathrm{mean}}$ are fixed normalized PCA projections. A lightweight two-layer MLP router $f_r$ produces two four-way categorical distributions
\begin{equation}
(p^{v},p^{s})
=\operatorname{softmax}\bigl(f_r(x)\bigr),
\end{equation}
where $p^{v}$ and $p^{s}$ denote the distributions over detail specificity and evidence scope, respectively. The softmax is applied independently within each head, and both heads are trained offline with cross-entropy supervision.

We combine the predictions into a joint distribution $p=p^{v}\otimes p^{s}$, where $\otimes$ denotes the Kronecker product. A rule-based scoring table $A$ maps the distribution $p$ to scores over candidate configurations. During inference, we select
\begin{equation}
(R,\rho,\psi)
=
\underset{c\in\mathcal{C}}{\arg\max}\;
p^{\top}A_c,
\end{equation}
where $\mathcal{C}$ is the set of candidate configurations and $A_c$ is the column of $A$ corresponding to $c$. The selected configuration specifies the resolution cap $R$, token retention ratio $\rho$, and wavelet setting $\psi$. The router and scoring table are shared across datasets and backbones. 

\subsection{Holistic Geographic Evidence Modeling}
We modulate token importance by jointly considering task relevance, spatial structure, and local token redundancy.

\noindent\textbf{Task-conditioned semantic evidence.}
UHR RSIs often contain diverse geographic objects over extensive areas, whereas a query may require evidence from only a small subset. Token importance is thus task-dependent, motivating task-conditioned semantic relevance as our first criterion. Since attention outputs decoupled from residual connections can better preserve fine-grained visual semantics \cite{lan2024clearclip} and facilitate text-guided localization \cite{ma2026desap}, we decompose the last-layer visual tokens $V^l$ into a residual component $V_d$ and an attention component $A_d$:
\begin{equation}
V^l = V_d + A_d + ffn\big(ln(V_d + A_d)\big), 
\end{equation}
where $V_d=V^{l-1}$ denotes the residual component from the previous layer, and $A_d$ denotes the projected self-attention:
\begin{equation}
A_d=proj\big(softmax(\frac{Q_lK_l^\top}{\sqrt{d_h}})V_l\big).
\end{equation}
Here, $ln$ and $ffn$ denote layer normalization and the feed-forward network, respectively. $Q_l$, $K_l$, and $V_l$ are the visual query, key, and value matrices. $proj$ is the attention output projection and $d_h$ is the attention-head dimension.

The localized semantics captured by $A_d$ provide a basis for identifying visual evidence relevant to the query. We quantify relevance after the backbone's multimodal projection, where $a_i,q_j\in\mathbb{R}^{e}$ denote the aligned attention features and query token representations, respectively. The task-relevance score of the $i$-th visual token is then computed as its mean cosine similarity with the query tokens:
\begin{equation}
E_t^{i}
=
\frac{1}{M}\sum_{j=1}^{M}
\frac{a_i^\top q_j}{\|a_i\|_2\|q_j\|_2},
\quad i=1,\ldots,N.
\end{equation}
A higher $E_{t}^{i}$ indicates stronger alignment between the visual token and the query, providing a task-aware semantic prior for token pruning. Such a prior is particularly effective in localizing key targets amid complex remote sensing background clutter, preventing pruning from being dominated by visually salient yet task-irrelevant regions.

\noindent\textbf{Scale-adaptive structure evidence.}
Beyond task relevance, structural cues in UHR RSIs encode critical geographic information, such as spatial connectivity and land-cover transitions. High-level attention representations in MLLMs emphasize semantic content but may not fully preserve fine-scale geographic structures \cite{lee2025frequency}. We therefore use wavelet decomposition \cite{vonesch2007generalized} to extract localized, multi-scale structural cues directly from the image, providing a complementary prior independent of semantic attention.

Given an input grayscale image $I\in\mathbb{R}^{H\times W}$, we first map it to the patch-level resolution with normalization, followed by an $L$-level discrete wavelet transform (DWT):
\begin{equation}
\{I^{\ell}_{LL}, I^{\ell}_{LH}, I^{\ell}_{HL}, I^{\ell}_{HH}\}_{\ell=1}^{L}=\operatorname{DWT}^{1:L}\big(\mathcal{N}(\mathcal{R}_{p}(I))\big),
\end{equation}
where $\mathcal{R}_{p}(\cdot)$ maps the image to the patch-level resolution and $\mathcal{N}(\cdot)$ denotes normalization. At each level $\ell$, $I^{\ell}_{LL}$ captures coarse appearance information, while $I^{\ell}_{LH}$, $I^{\ell}_{HL}$, and $I^{\ell}_{HH}$ encode horizontal, vertical, and diagonal structural variations. Since structure cues are mainly reflected in high-frequency responses, we aggregate the high-frequency subbands across scales to obtain token-level structural scores:
\begin{equation}
E_s =
\sum_{\ell=1}^{L}
\mathcal{U}_{t}(
\sum_{s\in\{\mathrm{LH},\mathrm{HL},\mathrm{HH}\}}
\left| I^{\ell}_{s} \right|
), 
\end{equation}
where $|\cdot|$ denotes the element-wise absolute value, $\mathcal{U}_{t}(\cdot)$ projects responses onto the final token grid, and we denote the resulting structural score of the $i$-th token by $E_s^i=[E_s]_i$. By modeling spatial structural score, tokens around boundaries, structures, and geographic transitions receive higher importance. Moreover, our framework adapts the decomposition strategy and level to the selected resolution, enabling scale-adaptive structural modeling.

\noindent\textbf{Dynamic local redundancy evidence.}
UHR RSIs contain extensive homogeneous regions, such as water bodies, farmland, and forests, where adjacent tokens may encode largely overlapping information. Semantic and structural scores assess individual token importance, but extended targets and repetitive textures may yield high-scoring yet redundant tokens. To obtain a compact, informative token set for downstream tasks, we introduce dynamic local redundancy evidence by modeling intrinsic local homogeneity and feature overlap among high-priority candidates.

Let $v_i$ denote the final-layer visual feature at position $p_i$, and let $\mathcal{L}_i$ contain its $3\times3$ neighbors. We quantify local homogeneity through Gaussian-weighted discrepancies:
\begin{equation}
E_d^i =
\frac{
\sum_{j\in\mathcal{L}_i}
a_{ij}\|v_i-v_j\|_2^2
}{
\sigma_v^2\sum_{j\in\mathcal{L}_i}a_{ij}
}, 
a_{ij}=e^{-\eta\|p_i-p_j\|_2^2},
\end{equation}
where $\sigma_v^2$ denotes feature variance and $\eta=0.5$ controls spatial weighting.
Lower $E_d^i$ indicates stronger local homogeneity.
Since local feature variation depends on land-cover composition, we further calibrate homogeneity using quartile statistics \cite{boukerche2020outlier}. Let $Q_1$ and $Q_3$ denote the first and third quartiles of the scores ${E_d^i}$, with $\mathrm{IQR}=Q_3-Q_1$. The spatial homogeneity score is defined as
\begin{equation}
R_i = (Q_3-E_d^i) / \mathrm{IQR}.
\end{equation}
We restrict $R_i$ to $[0,1]$,  with larger values indicating stronger homogeneity. Since homogeneity alone does not capture representational overlap, we further quantify feature affinity to nearby high-priority candidates:
\begin{equation}
P_i =
\max_{j\in\mathcal{C}_i}
\left[\operatorname{cos}(z_i,z_j)-b\right]_{+},
\end{equation}
where $z_i$ is the projected embedding, $\mathcal{C}_i$ contains nearby higher-priority candidates, $b$ is the median local cosine similarity, and $[x]_{+}=\max(x,0)$. The initial candidate pool comprises the top-$k$ tokens ranked by $S_i$ with $P_i=0$. We combine spatial homogeneity and feature affinity into the dynamic local redundancy evidence, $E_r^i=0.25R_i+P_i$. It complements semantic and structural importance by reducing repeated selection from large targets or repetitive textures, yielding a compact and informative representation.
\begin{table*}[t]
\centering\scriptsize\color{black}
\setlength{\tabcolsep}{2.2pt}
\resizebox{\textwidth}{!}{%
\begin{tabular}{@{}lcc|rrrrrrrrr|rrrr|rrrr@{}}
\toprule
\multirow{2}{*}{\textbf{Method}} & \multirow{2}{*}{\shortstack{\textbf{Max}\\\textbf{Size}}} & \multirow{2}{*}{\shortstack{\textbf{Avg.}\\\textbf{Tok.}}} & \multicolumn{9}{c|}{\textbf{XLRS-Bench}} & \multicolumn{4}{c|}{\textbf{LRS-VQA}} & \multicolumn{4}{c}{\textbf{MME-RealWorld-RS}} \\
\cmidrule(lr){4-12}\cmidrule(lr){13-16}\cmidrule(lr){17-20}
&&& Cnt & SC & OSR & OP & Plan & AR & CR & SR & Avg. & FAIR & Bridge & STAR & Avg. & Color & Count & Pos. & Avg. \\
\midrule
\rowcolor{sectionbg}\multicolumn{20}{l}{\textit{Closed-source VLMs}} \\
GPT-4o & -- & -- & 29.4 & 49.0 & 24.6 & 10.4 & 35.0 & 73.0 & 49.0 & 20.0 & 23.0 & 22.2 & 31.8 & 27.4 & 26.4 & 34.2 & 15.2 & 37.1 & 28.9 \\
Claude 3.7 Sonnet & -- & -- & 24.4 & 51.3 & 27.6 & 31.4 & 35.0 & 65.0 & 55.5 & 28.3 & 35.1 & 14.3 & 28.5 & 15.0 & 16.7 & 32.9 & 19.4 & 28.6 & 27.0 \\
GPT-4o-mini & -- & -- & 24.4 & 46.0 & 23.6 & 37.0 & 29.0 & 71.0 & 50.5 & 6.7 & 36.2 & 18.7 & 32.0 & 25.9 & 24.6 & 5.8 & 2.6 & 11.5 & 6.7 \\
\midrule
\rowcolor{sectionbg}\multicolumn{20}{l}{\textit{Open-source General VLMs}} \\
LLaVA-v1.5 & $336^{2}$ & 576 & 22.5 & 19.7 & 23.0 & 20.2 & 39.0 & 35.0 & 34.0 & 30.0 & 22.9 & 18.8 & 30.7 & 22.6 & 22.6 & 23.1 & 16.9 & 26.3 & 22.1 \\
LLaVA-v1.6 & $672^{2}$ & 2,880 & 35.0 & 46.3 & 30.0 & 32.1 & 27.0 & 69.0 & 57.0 & 35.0 & 36.0 & 19.6 & 32.0 & 23.6 & 23.6 & 24.5 & 18.0 & 27.5 & 23.4 \\
LLaVA-OV & $2,304^{2}$ & 8,667 & 33.1 & 49.0 & 25.2 & 36.7 & 24.0 & \underline{76.0} & 59.5 & 43.3 & 38.3 & 20.6 & 35.1 & 26.1 & 25.7 & 26.8 & 26.1 & 27.6 & 26.8 \\
IXC-2.5 & $2,800^{2}$ & 10,400 & 31.9 & 37.7 & 26.0 & 22.1 & 41.0 & 72.0 & 64.0 & 36.7 & 30.0 & 25.3 & \textbf{38.4} & 27.3 & 28.3 & 45.3 & 17.6 & 45.0 & 36.1 \\
Qwen2.5-VL-7B & $3,584^{2}$ & 16,384 & 37.5 & 61.7 & 36.2 & 41.5 & 27.0 & 68.0 & 58.5 & 38.3 & 43.8 & 22.9 & 36.9 & 25.8 & 26.5 & 45.9 & 21.1 & 55.2 & 40.9 \\
Qwen3-VL-8B & $4,096^{2}$ & 16,384 & 38.1 & 63.3 & 30.6 & 45.4 & 37.0 & 75.0 & 64.5 & 43.3 & 46.3 & 24.6 & \underline{38.2} & 27.5 & 28.2 & 48.1 & 21.3 & 56.4 & 42.1  \\
\midrule
\rowcolor{sectionbg}\multicolumn{20}{l}{\textit{Remote Sensing VLMs}} \\
GeoChat & $504^{2}$ & 1,296 & 24.4 & 16.0 & 24.2 & 19.5 & 10.0 & 33.0 & 32.0 & 24.0 & 21.2 & 20.2 & 24.5 & 13.8 & 17.3 & 23.1 & 15.7 & 25.0 & 21.3 \\
EarthDial & $2,833^{2}$ & 10,496 & 33.1 & 24.3 & 24.8 & 32.4 & 43.0 & 62.0 & 60.5 & 48.3 & 33.9 & 17.3 & 24.5& 17.0 & 18.2 & 35.1 & 22.5 & 44.1 & 34.0 \\
VHM & $336^{2}$ & 576 & 27.5 & 27.3 & 33.0 & 26.8 & 39.0 & 53.0 & 46.0 & 26.7 & 30.4 & 22.6 & 34.4 & 26.8 & 26.6 & 20.3 & 16.8 & 35.2 & 24.2 \\
\midrule
\rowcolor{sectionbg}\multicolumn{20}{l}{\textit{Direct UHR Perception Methods}} \\
GeoLLaVA-8K & $8,064^{2}$ & 13,824 & 33.8 & 62.3 & 35.0 & 37.6 & \textbf{66.0} & 67.0 & 65.0 & 50.0 & 43.3 & 20.8& 32.7& 24.6 &24.6& 27.9& 22.3 &34.9& 28.4 \\
VisionZip & $8,064^{2}$ & 13,824 & 33.1 & 45.0 & 38.6 & 38.1 & \underline{62.0} & 36.0 & 42.5 & 46.7 & 39.8 & 19.1 & 30.0 & 22.5 & 22.5 & 25.7&20.5	&32.1&26.1 \\
SparseVLM & $8,064^{2}$ & 13,824 & 32.5 & 34.3 & \underline{38.8} & 31.4 & \underline{62.0} & 43.0 & 45.0 & 45.0 & 35.5 & 17.0 & 26.8& 20.1 & 20.1 & 22.9&18.3&28.7&23.3 \\
AdaGEM$^{\dagger}$ (Ours) & $8,064^{2}$ & 13,824 & 35.6 & 61.7 & 35.4 & 39.5 & \textbf{66.0} & 66.0 & 63.0 & \underline{51.7} &44.3& 22.0& 33.5 & 25.5 & 25.6 &28.8&22.8&35.6&29.1	 \\
\rowcolor{oursbg}
\textbf{AdaGEM$^{\dagger}$ (Ours)} & Dynamic & 9,858 & 36.9 & \underline{63.7} & 36.4 & 39.5 & \textbf{66.0} & 66.0 & 64.0 & \textbf{53.3} & 44.8  & 23.2&34.0& 26.0& 26.3 & 29.5&23.2&36.2&29.7 \\
\rowcolor{oursbg}
\multicolumn{3}{@{}c|}{\hspace{0.6em}\gain{\textit{$\Delta$ vs.\ GeoLLaVA-8K backbone}}}
& \gain{+3.1} & \gain{+1.4} & \gain{+1.4} & \gain{+1.9}
& \gain{0.0} & \gain{-1.0} & \gain{-1.0} & \gain{+3.3} & \gain{+1.5}
& \gain{+2.4} & \gain{+1.3} & \gain{+1.4} & \gain{+1.7}
& \gain{+1.6} & \gain{+0.9} & \gain{+1.3} & \gain{+1.3} \\
\midrule
\rowcolor{sectionbg}\multicolumn{20}{l}{\textit{Multi-scale Perception Methods}} \\
UHR-BAT &  $4,032^{2}$ & 11,100 & 28.8 & 53.7 & \textbf{44.8} & 39.6 & 54.0 & 62.0 & 61.0 & 46.7 & 44.0 & 16.6 & 23.5 & 20.0 & 19.5 & 42.0 & 14.0 & 44.0 & 33.5 \\
ZoomSearch & Dynamic & 8,667 & \textbf{43.8} & 51.0 & 32.2 & 41.4 & 29.0 & 74.0 & 59.5 & 43.3 & 42.8 & 21.8 & 32.3 & 26.1 & 25.7 & 48.6 & 20.1 & 59.8 & 43.0 \\
ZoomEarth & Dynamic & 2,738 & 28.7 & 51.0 & 20.8 & 35.7 & 38.0 & 70.0 & 52.5 & 28.3 & 36.5 & 20.0 & 23.2 & 21.6 & 21.3 & 41.9 & 27.6 & 46.6 & 38.8 \\
RFM-DIP & Dynamic & 3,456 & \underline{39.4} & 37.0 & 25.6 & 34.6 & 24.0 & 70.0 & 60.0 & \textbf{53.3} & 36.4 & 21.9 & \underline{38.2} & 26.7 & 26.9 & 44.7 & \textbf{31.0} & 49.7 & 41.9 \\
WeaveEarth & Dynamic & 4,551 & 37.5 & 50.3 & 36.4 & \underline{47.4} & 36.0 & 73.0 & \textbf{67.5} & 46.7 & \underline{47.1} & \underline{31.0} & 26.1 & \underline{36.7} & \underline{33.4} & \underline{51.4} & \underline{27.7} & \underline{62.6} & \underline{47.4} \\
\rowcolor{oursbg}
\textbf{AdaGEM (Ours)} & Dynamic & 2,218 & 38.1 & \textbf{65.3} & 29.0 & \textbf{47.9}
& 43.0  & \textbf{77.0} & \underline{67.0} & 48.3 & \textbf{48.1}  & \textbf{31.5} & 34.8 & \textbf{37.1} & \textbf{35.0} & \textbf{64.2} & 20.4 & \textbf{65.2} & \textbf{50.2} \\
\rowcolor{oursbg}
\multicolumn{3}{@{}c|}{\hspace{0.6em}\gain{\textit{$\Delta$ vs.\ Qwen3-VL-8B backbone}}}
& \gain{0.0} & \gain{+2.0} & \gain{-1.6} & \gain{+2.5}
& \gain{+6.0} & \gain{+2.0} & \gain{+2.5} & \gain{+5.0} & \gain{+1.8}
& \gain{+6.9} & \gain{-3.4} & \gain{+9.6} & \gain{+6.8}
& \gain{+16.1} & \gain{-0.9} & \gain{+8.8} & \gain{+8.1} \\
\bottomrule
\end{tabular}}
\caption{\textbf{Results on XLRS-Bench, LRS-VQA, and MME-RealWorld-RS.} `Avg.' denotes weighted average sub-task accuracy. Best and second-best results are \textbf{bolded} and \underline{underlined}.}
\vspace{-2mm}
\label{tab:all_benchmarks}
\end{table*}
\subsection{Evidence Modulation for Inference}  
We integrate task relevance $E_t^i$, structural importance $E_s^i$, and dynamic local redundancy evidence $E_r^i$ to construct a compact, informative visual representation. We couple semantic and structural evidence through multiplication so that structural cues guide scene-wide selection when queries without specific targets yield similar positive semantic scores, whereas target-specific queries favor local structures and semantically related context. To promote complementary evidence, a weighted redundancy penalty further reduces the priority of repetitive tokens. The resulting importance score is defined as
\begin{equation}
S_i =
R\left(E_t^i E_s^i\right)
-\lambda E_r^i,
\end{equation}
where $R$ denotes normalized ranking, and $\lambda=0.25$ controls the local redundancy penalty. For Qwen3-VL, pruning is applied after spatial merging. We retain the top-$k$ tokens ranked by $S_i$ while preserving their original projected features and spatial order. The retained tokens are then combined with the query tokens for LLM inference.

\section{Experiments}
\subsection{Experimental Settings}
\noindent\textbf{Datasets.}
We evaluate our pruning framework on three complementary remote sensing VQA benchmarks: XLRS-Bench \cite{wang2025xlrsbench}, LRS-VQA \cite{LRS-VQA}, and MME-RealWorld-RS \cite{zhang2025mmerealworld}. MME-RealWorld-RS \cite{zhang2025mmerealworld} provides expert-annotated multiple-choice questions on HR-RSIs, covering color recognition, object counting, and position understanding. LRS-VQA \cite{LRS-VQA} evaluates visual perception of large RSIs through 7,333 open-ended questions spanning 8 question types. XLRS-Bench \cite{wang2025xlrsbench} evaluates perception and reasoning in UHR RSIs across 13 categories on 3,080 questions.

\noindent\textbf{Baselines.}
We compare with closed-source \cite{hurst2024gpt4o,anthropic2025claude37}, open-source general-purpose \cite{liu2024llava,li2024llavaonevisioneasyvisualtask,bai2025qwen25vltechnicalreport}, and RS-specific VLMs \cite{kuckreja2023geochat,EarthDial,pang2025vhm}. Direct UHR baselines include GeoLLaVA-\textit{8K} \cite{wang2025geollava8k}, alongside VisionZip \cite{yang2025visionzip} and SparseVLM \cite{zhang2024sparsevlm} implemented on this backbone. Multi-scale UHR baselines include UHR-BAT \cite{dang2026uhrbat}, ZoomSearch \cite{ZoomSearch}, ZoomEarth \cite{zoomearth}, RFM-DIP \cite{LRS-VQA}, and WeaveEarth \cite{WeaveEarth}. To assess generality across MLLM backbones, we implement AdaGEM$^{\dagger}$ on LongVA-based \cite{zhang2024longva} GeoLLaVA-8K and AdaGEM on the same Qwen3-VL-8B \cite{bai2025qwen3vl} backbone as WeaveEarth. 

\subsection{Main Results}
\noindent\textbf{Direct UHR perception.}
We first compare AdaGEM with direct UHR perception methods, which encode the full scene at UHR and aggressively prune the resulting visual tokens. Using GeoLLaVA-8K \cite{wang2025geollava8k} as a common frozen backbone, we replace its original token pruning module with VisionZip \cite{yang2025visionzip}, SparseVLM \cite{zhang2024sparsevlm}, and our geospatial evidence-guided token selection module. To isolate the contribution of token selection, we evaluate these alternatives under the same resolution cap of $8{,}064 \times 8{,}064$ and a token retention rate of 4.17\%. As shown in Table~\ref{tab:all_benchmarks}, AdaGEM$^{\dagger}$ outperforms GeoLLaVA-8K and both generic pruning baselines across all three benchmarks under this fixed configuration, while incorporating the question-conditioned router further improves accuracy, yielding overall gains of 1.5\%, 1.7\%, and 1.3\% over GeoLLaVA-8K on XLRS-Bench, LRS-VQA, and MME-RealWorld-RS, respectively, while reducing the reported visual-token count from 13,824 to 9,858 (28.7\%). These results support the value of geospatial evidence modulation for token selection under a fixed visual budget, with question-conditioned granularity adaptation enabling further accuracy gains at lower token cost.

\noindent\textbf{Multi-scale perception.} 
We further compare AdaGEM with UHR-BAT \cite{dang2026uhrbat}, ZoomSearch \cite{ZoomSearch}, ZoomEarth \cite{zoomearth}, RFM-DIP \cite{LRS-VQA}, and WeaveEarth \cite{WeaveEarth}, which organize visual evidence through multi-scale processing or adaptive region selection.
We report the mean number of visual tokens fed to the LLM across all evaluation datasets.
Table~\ref{tab:all_benchmarks} shows that AdaGEM uses the fewest visual tokens among these methods and achieves the highest overall accuracy on all three benchmarks.
Compared with WeaveEarth, which uses the same Qwen3-VL-8B backbone, AdaGEM uses 51.3\% fewer visual tokens and improves accuracy by 1.0\%, 1.6\%, and 2.8\% on XLRS-Bench, LRS-VQA, and MME-RealWorld-RS, respectively.
The largest gain occurs on MME-RealWorld-RS, where AdaGEM achieves 50.2\% overall accuracy, with strong results in color and position recognition.
These gains extend across benchmarks with different image scales and task compositions.
Together, these results support aligning visual granularity and evidence retention with question requirements for more accurate UHR RSI understanding with fewer visual tokens.

\begin{figure*}[t]
\centering
\includegraphics[width=0.99\textwidth]{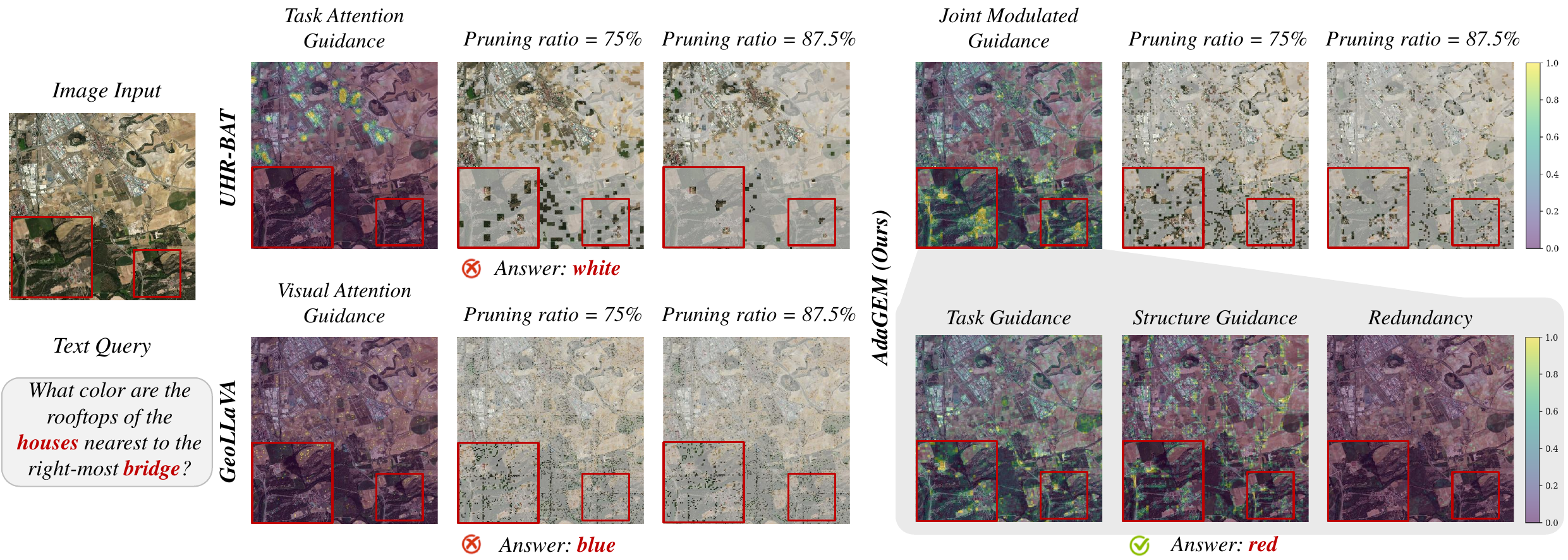}
\vspace{-3mm}
\caption{\textbf{Visualization analysis on XLRS-Bench comparing GeoLLaVA-$\textit{8K}$ \cite{wang2025geollava8k}, UHR-BAT \cite{dang2026uhrbat}, and AdaGEM (Ours).} The original image and the corresponding pruning guidance with pruned results at pruning ratios of 75\% and 87.5\%  are shown. Gray areas indicate discarded tokens, while red boxes highlight the query-relevant regions.} \label{fig_exp}
\vspace{-2mm}
\end{figure*}
\noindent\textbf{Qualitative comparisons.}
We visualize the patches retained at pruning ratios of 75\% and 87.5\% on XLRS-Bench \cite{wang2025xlrsbench} at $4096 \times 4096$, comparing GeoLLaVA-$\textit{8K}$ \cite{wang2025geollava8k}, UHR-BAT \cite{dang2026uhrbat}, and AdaGEM alongside criterion-wise activation maps. As shown in Figure~\ref{fig_exp}, UHR-BAT and GeoLLaVA-\textit{8K} discard query-relevant regions and produce incorrect answers. AdaGEM combines task relevance, structural cues, and local redundancy suppression to retain the queried targets and their spatial context, yielding correct answers. These observations provide qualitative support for preserving geographic evidence through complementary guidance under aggressive pruning.

\subsection{Ablation Studies}

\noindent\textbf{Component ablation.}
We examine the complementary roles of granularity routing and evidence guidance in Table~\ref{tab:ablation_routing_evidence}. Without routing, we fix the resolution cap at $8{,}064$ and the token retention ratio at 4.17\%. Disabling evidence guidance retains GeoLLaVA's native region-clustering compression. Under the fixed configuration, evidence guidance improves overall accuracy from 43.3\% to 44.3\%, demonstrating its contribution to token selection at a controlled budget. Adding routing further improves accuracy to 44.8\% while reducing average visual-token usage from 13,824 to 9,858 (28.7\%). The full model achieves the highest perception and overall accuracy, although its reasoning accuracy remains slightly below the native baseline. Routing alone yields a marginal 0.1\% accuracy gain on GeoLLaVA-8K, potentially constrained by its reliance on visual saliency for evidence selection. Complementary evidence guidance increases this gain to 0.5\%, supporting the joint adaptation of visual granularity and evidence selection.
\begin{table}[t]
\centering
\begingroup
\small
\setlength{\tabcolsep}{3pt}
\renewcommand{\arraystretch}{1.16}
\makeatletter
\newcommand{\fullrowbg}[1]{%
  \rlap{\color{#1}%
    \rule[-\dp\@arstrutbox]{0.90\linewidth}%
    {\dimexpr\ht\@arstrutbox+\dp\@arstrutbox\relax}}%
}
\makeatother
\begin{tabular*}{0.9\linewidth}{@{\extracolsep{\fill}}lrrrr@{}}
\toprule
\textbf{Variant} & \textbf{Perception} & \textbf{Reasoning} & \textbf{Avg.}$\uparrow$ & \textbf{\boldmath$\Delta$} \\
\midrule
Native baseline & 39.7 & 63.7 & 43.3 & --- \\
Routing only & 39.8 & \textbf{63.9} & 43.4 & +0.1 \\
Evidence only & 41.0 & 62.8 & 44.3 & +1.0 \\
\fullrowbg{lightblue!60}\textbf{AdaGEM$^{\dagger}$} & \textbf{41.5} & 63.5 & \textbf{44.8} & \textbf{+1.5} \\
\bottomrule
\end{tabular*}
\endgroup
\caption{\textbf{Component ablation on XLRS-Bench.}
Avg. denotes overall accuracy (\%). $\Delta$ denotes gain over the native baseline.}
\vspace{-4mm}
\label{tab:ablation_routing_evidence}
\end{table}

\begin{figure}[t]
    \centering
    \includegraphics[width=\linewidth]{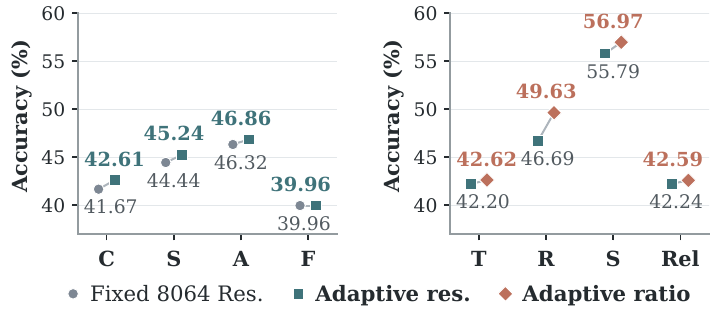}
    \vspace{-4mm}
    \caption{\textbf{Adaptive configurations on XLRS-Bench.}
\textit{(Left)} Accuracy across detail groups with adaptive $R$ at fixed $\rho,\psi$ , and \textit{(Right)} across evidence scopes when adapting $\rho$ at fixed $R,\psi$.}
    \label{fig:adaptive_configuration}
    \vspace{-4mm}
\end{figure}

\noindent\textbf{Effectiveness of granularity routing.}
We evaluate the downstream benefits of adapting resolution and token retention to question requirements. The router is trained on questions from datasets distinct from the evaluation benchmarks \cite{li2024vrsbench, rshr_benchmark, wang2025geollava8k}, and the same router and scoring table are shared across datasets and backbones. As shown in Figure \ref{fig:adaptive_configuration}, adaptive resolution improves accuracy over the fixed $8064^2$ resolution by 0.95\%, 0.79\%, and 0.54\% in the three lower-resolution groups, respectively. Adaptive token retention improves accuracy across all four evidence scopes, particularly for regional and scene-wide evidence, with gains of 2.94\% and 1.19\%, respectively. These results support adapting resolution to the required level of detail and token retention to evidence scope, enabling more effective visual allocation for different questions. 

\noindent\textbf{Effectiveness of geographic evidence.}
We ablate AdaGEM's task relevance, spatial structure, and local redundancy criteria on XLRS-Bench under the same adaptive router. As shown in Table \ref{tab:ablation_guidance}, task relevance achieves the strongest single-criterion overall accuracy of 47.1\%, whereas structure and redundancy alone yield 45.4\% and 45.3\%, respectively. Combining task relevance with structure or redundancy suppression increases overall accuracy to 47.8\% and 47.6\%. Specifically, adding structure improves perception accuracy from 44.8\% to 45.6\%, while adding redundancy suppression to the task-and-structure combination raises reasoning accuracy from 60.1\% to 61.5\%. Integrating all three criteria achieves the best perception, reasoning, and overall accuracies of 45.7\%, 61.5\%, and 48.1\%, respectively, supporting their complementary roles in geographic evidence selection.

\begin{table}[t]
\centering
\renewcommand{\arraystretch}{1.15}
\resizebox{\columnwidth}{!}{%
\begin{tabular}{ccc|ccc}
\toprule
\textbf{Task} & \textbf{Structure} & \textbf{Redundancy}
& \textbf{Perception} & \textbf{Reasoning} & \textbf{Avg.} \\
\midrule

\rowcolor{gray!10}\multicolumn{3}{c|}{\textit{\textcolor{gray}{Qwen3-VL-8B backbone}}} & \textcolor{gray}{44.2} & \textcolor{gray}{58.0} & \textcolor{gray}{46.3} \\

\hline
\cmark & \xmark & \xmark
& 44.8 & 59.8 & 47.1 \\

\xmark & \cmark & \xmark
& 43.2 & 57.8  & 45.4 \\

\xmark & \xmark & \cmark
& 42.9 & 58.5 & 45.3 \\

\cmark & \xmark & \cmark
& 45.2 & \underline{61.1} & 47.6  \\

\xmark & \cmark & \cmark
& 43.5 & 57.6& 45.6 \\

\cmark & \cmark & \xmark
 & \underline{45.6} & 60.1 & \underline{47.8} \\

\rowcolor{lightblue!60}
\cmark & \cmark & \cmark &
\textbf{45.7} & \textbf{61.5} & \textbf{48.1}  \\

\bottomrule
\end{tabular}%
}
\caption{\textbf{Evidence ablation of AdaGEM on XLRS-Bench.}
The first gray row is the unpruned baseline with a $4096^2$ resolution. All pruned variants share routing and token budgets.}
\label{tab:ablation_guidance}
\vspace{-2mm}
\end{table}

\begin{figure}[t]
\centering
\includegraphics[width=\columnwidth]{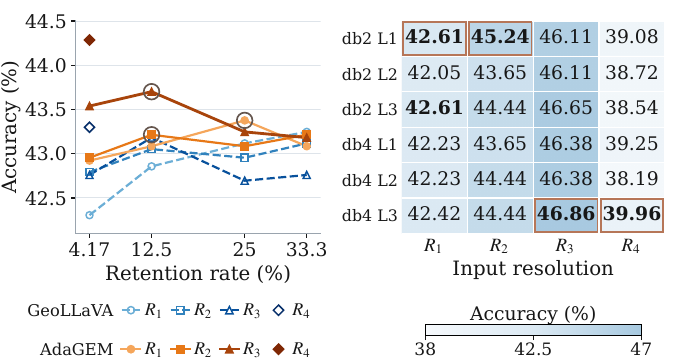}
\vspace{-5mm}
\caption{\textbf{Retention and wavelet sensitivity on XLRS-Bench.}
\textit{(Left)} Overall accuracy trends across retention ratios. Circles mark AdaGEM$^{\dagger}$'s peaks.
\textit{(Right)} Wavelet ablation within router-selected groups, keeping resolution and retention fixed.
} 
\label{fig:wavelet_resolution}
\vspace{-2mm}
\end{figure}

\begin{figure}[t]
    \centering
    \includegraphics[width=0.86\linewidth]{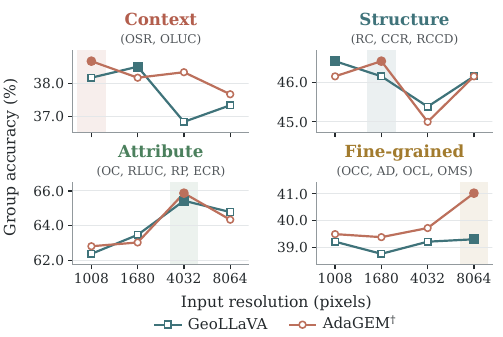}
    \vspace{-3mm}
\caption{\textbf{Task-dependent granularity preferences on XLRS-Bench.}
Accuracy across four L3 subtask groups with OA-selected retention at each resolution. Shading marks AdaGEM$^{\dagger}$'s peaks.}
    \label{fig:ratio_reslolution}
    \vspace{-2mm}
\end{figure}

\noindent\textbf{Sensitivity to granularity configurations.}
On XLRS-Bench, we examine token retention and wavelet settings for AdaGEM$^{\dagger}$ across four resolution caps: 1008, 1680, 4032, and 8064 pixels.
With db4 level~3 fixed, the best tested retention ratio, selected by overall benchmark accuracy, shifts from 25\% at $R=1008$ to 12.5\% at $R=4032$ (Figure \ref{fig:wavelet_resolution}, \textit{Left}).
At $R=4032$, increasing retention to 33.3\% reduces accuracy from 43.70\% to 43.18\%, showing that more tokens do not necessarily improve performance.
Only 4.17\% retention is evaluated at $R=8064$.
Within frozen granularity groups, with each query's resolution and retention fixed, db2 level~1 is best or tied for best at lower resolutions, whereas db4 level~3 performs best at higher resolutions (Figure \ref{fig:wavelet_resolution}, \textit{Right}).
Using the selected retention ratio at each resolution for both methods, with db4 level~3 fixed for AdaGEM$^{\dagger}$, we then examine resolution sensitivity across four L3 subtask groups: Context, Structure, Attribute, and Fine-grained.
Figure \ref{fig:ratio_reslolution} shows distinct granularity preferences, with some groups peaking below the maximum resolution.
These results motivate jointly adapting resolution and token retention to task requirements.

\begin{table}[t]
\centering
\setlength{\tabcolsep}{3.1pt}
\renewcommand{\arraystretch}{1.15}
\resizebox{\columnwidth}{!}{%
\begin{tabular}{@{}lrrrrrr@{}}
\toprule
\textbf{Method} & \textbf{V-Tokens}$\downarrow$
& $T_{\mathrm{enc}}\downarrow$
& $T_{\mathrm{pf}}\downarrow$ & $T_{\mathrm{total}}\downarrow$
& \textbf{TFLOPs}$\downarrow$ & \textbf{Acc.}$\uparrow$ \\
\midrule
\rowcolor{gray!10}\multicolumn{7}{l}{\textit{GeoLLaVA backbone}} \\
GeoLLaVA-8K & 13,824
& 1.17 {\footnotesize ($1.0\times$)}
& 0.88 {\footnotesize ($1.0\times$)}
& 1.86 {\footnotesize ($1.0\times$)}
& 192.83 & 43.3 \\
AdaGEM$^{\dagger}$-8K & 13,824
& 1.28 {\footnotesize ($0.9\times$)}
& 0.88 {\footnotesize ($1.0\times$)}
& 1.86 {\footnotesize ($1.0\times$)}
& 192.83 & 44.3 \\
\rowcolor{lightblue!60}
AdaGEM$^{\dagger}$ & 9,858
& 0.53 {\footnotesize ($2.2\times$)}
& 0.78 {\footnotesize ($1.1\times$)}
& 1.77 {\footnotesize ($1.1\times$)}
& 167.52 & 44.8 \\
\midrule
\rowcolor{gray!10}\multicolumn{7}{l}{\textit{Qwen3-VL-8B backbone}} \\
Qwen3-VL-8B & 16,358
& 4.32 {\footnotesize ($1.0\times$)}
& 1.56 {\footnotesize ($1.0\times$)}
& 2.48 {\footnotesize ($1.0\times$)}
& 308.86 & 46.3 \\
WeaveEarth & 4,029
& 3.37 {\footnotesize ($1.3\times$)}
& 0.38 {\footnotesize ($4.1\times$)}
& 1.35 {\footnotesize ($1.8\times$)}
& 71.54 & 47.1 \\
AdaGEM-4K & 2,043
& 4.32 {\footnotesize ($1.0\times$)}
& 0.20 {\footnotesize ($7.8\times$)}
& 1.13 {\footnotesize ($2.2\times$)}
& 31.86 & 46.8 \\ 
\rowcolor{lightblue!60}
\textbf{AdaGEM} & \textbf{2,014}
& \textbf{3.52 {\footnotesize ($1.2\times$)}}
& \textbf{0.19 {\footnotesize ($8.2\times$)}}
& \textbf{1.09 {\footnotesize ($2.3\times$)}}
& \textbf{31.36} & \textbf{48.1} \\
\bottomrule
\end{tabular}}
\caption{\textbf{Efficiency comparisons on XLRS-Bench.}
We report average token counts, visual encoding time (s),
LLM prefill time (s), total LLM inference time (s),
FLOPs (T), and accuracy (\%).}
\label{tab:efficiency}
\end{table}

\subsection{Efficiency Analysis}
Table~\ref{tab:efficiency} compares inference efficiency on XLRS-Bench. $T_{\mathrm{enc}}$, $T_{\mathrm{pf}}$, and $T_{\mathrm{total}}$ denote visual encoding latency, LLM prefill latency, and total LLM inference latency including prefill and decoding, respectively. The LLM timings exclude visual encoding, routing, and pruning. Compared with Qwen3-VL-8B, AdaGEM reduces visual encoding latency by 18.5\% and achieves an $8.2\times$ prefill speedup and a $2.3\times$ total LLM speedup, while reducing analytical FLOPs by 89.85\% and improving accuracy from 46.3\% to 48.1\%. Relative to WeaveEarth, it uses 50.0\% fewer visual tokens and halves prefill latency. AdaGEM improves accuracy over its fixed 4K counterpart by 1.3\% while reducing visual encoding and LLM inference latency. On the GeoLLaVA backbone, AdaGEM$^{\dagger}$ reduces visual encoding latency from 1.28\,s to 0.53\,s relative to its fixed 8K counterpart, while improving accuracy from 44.3\% to 44.8\% and modestly reducing LLM latency. These results demonstrate AdaGEM's efficiency advantages across backbones.

\section{Conclusion}
In this work, we presented AdaGEM, a unified framework jointly adapting visual granularity to task demands and preserving essential geospatial evidence. A task-adaptive router configures input resolution and token retention, while an evidence modulator combines task relevance, wavelet-based structure, and local redundancy suppression to select complementary visual tokens. Experiments across remote sensing benchmarks and MLLM backbones demonstrate both higher accuracy and lower inference costs than compared methods for UHR remote sensing understanding.

\section*{LLM Usage Statement}
The authors used ChatGPT (OpenAI) to refine the wording and organization of author-written text, including the introduction, method descriptions, and experimental discussion, to improve grammar, clarity, and readability. ChatGPT was also used to refine the visual presentation of author-produced experimental plots, including colors, fonts, and layout. All AI-assisted revisions were reviewed and approved by the authors. The authors take full responsibility for the final content of this work, including its scientific claims, experimental results, and figures.

{
    \small
    \bibliographystyle{ieeenat_fullname}
    \bibliography{aaai2027}
}

\end{document}